\documentclass[letterpaper, 10 pt, conference]{ieeeconf}  % Comment this line out if you need a4paper

\usepackage{tikz}
\usepackage[utf8]{inputenc} % allow utf-8 input
\usepackage[T1]{fontenc}    % use 8-bit T1 fonts
\usepackage{hyperref}       % hyperlinks
\usepackage{url}            % simple URL typesetting
\usepackage{booktabs}       % professional-quality tables
\usepackage{amsfonts}       % blackboard math symbols
\usepackage{nicefrac}       % compact symbols for 1/2, etc.
\usepackage{microtype}      % microtypography
\usepackage{graphicx}
\usepackage{pgfplots}

\IEEEoverridecommandlockouts                              % This command is only needed if 
\title{\LARGE \bf
Improving benchmarks for autonomous vehicles testing using synthetically generated images
}

\author{Aleksander Lukashou. \\
Department of Computer Science \\
AGH University of Technology\\ 
Krakow, Poland \\
{\tt\small lukashou@agh.edu.pl}}
\begin{document}

\maketitle
\thispagestyle{empty}
\pagestyle{empty}

%%%%%%%%%%%%%%%%%%%%%%%%%%%%%%%%%%%%%%%%%%%%%%%%%%%%%%%%%%%%%%%%%%%%%%%%%%%%%%%%
\begin{abstract}

Nowadays autonomous technologies are a very heavily explored area and particularly computer vision
as the main component of vehicle perception. The quality of the whole vision system based on neural
networks relies on the dataset it was trained on. It’s extremely difficult to find traffic sign datasets
from most of the counties of the world. Meaning autonomous vehicle from the USA won’t be able to
drive though Lithuania recognizing all road signs on the way. In this paper, we propose a solution on how to update model using a small dataset from the country vehicle will be used in. It’s important
to mention that is not panacea, rather small upgrade which can boost autonomous car development
in countries with limited data access. We achieved about 10\% quality raise and expect even better
results during future experiments.

\end{abstract}

%%%%%%%%%%%%%%%%%%%%%%%%%%%%%%%%%%%%%%%%%%%%%%%%%%%%%%%%%%%%%%%%%%%%%%%%%%%%%%%%
\section{INTRODUCTION}
Driver-less cars wildly use techniques based on neural networks to detect and classify traffic signs. To attempt \textit{well-tuned state} such models need to be trained with a significant amount of data. In case of traffic signs that means photos taken in different weather and lightning conditions, actor speed etc. Many researchers working in autonomous car development are facing a lack of available qualitative datasets. Massive labeled datasets are publicly accessible for developed countries such as Belgium, Germany or China. However, for singe enthusiasts or private companies in smaller countries collecting and labeling data is too costly and time consumable. In this paper, we propose an alternative way to build a sizable benchmark for model training.

In order to prove our main idea that GAN\footnote{Generative Adversarial Network}-generated images of traffic signs perform better than classical data augmentation methods and artificial images synthesis allows us to achieve massive datasets, we have done following research milestones:
\begin{itemize}
\item Compiled road signs from Belgian Traffic Signs Dataset\cite{Timofte-WACV-2009}, The German Traffic Sign Benchmark\cite{Stallkamp2012}, TSRD - which is part of Chinese Traffic Sign Database. Down below we are going to use a reference to this dataset as Superset. Superset consists of about 25000 images. Test dataset contains randomly extracted images from Superset, Test dataset counts 2500 images. 
\item GAN has been built using commonly used architecture and so has been built CNN\footnote{Convolutional Neural Network}-classifier.
\item As the first step of our evaluation we are going to verify proficiency of Classifier which was pretrained on single GTSRB\footnote{German Traffic Sign Recognition Benchmark} by testing it Test dataset.
\item Classical methods of data augmentation can boost Classifier performance and we prove it by re-train neural network with GTSRB extended with images we receive as a result of augmentation. Compare testing of outcome model on the same Test dataset.
\item Next we are going to train our GAN and retrain Classifier on Superset extended with GAN-generated images.
\item Last step is to test our newly trained Classifier on the same Test dataset.
\item Later or we summarize conclusions and propose further improvement.
\end{itemize}
\textbf{Important note}: As traffic signs not precisely standardised and might differ crucially in different countries, we are going to use only those classes which consist similar essential features, more details provided in Table~\ref{tab:table}..

\begin{table}[h]
\caption{Used Traffic Sign Classes}
\label{table_example}
\begin{center}
\begin{tabular}{|c||c||c|}
\hline
    0     & Closed to all in both directions  & circle, white, red\\
\hline
    1     & No entry & circle, white stripe, red      \\
\hline
    2     & Stop and give way       & white text, red  \\
\hline
    3     & Speed limit 30  & circle, white, red, black text     \\
\hline
    4     & Speed limit 50 & circle, white, red, black text   \\
\hline    
    5     & Speed limit 70  & circle, white, red, black text   \\
\hline
    6     & Speed limit 100  & circle, white, red, black text  \\
\hline
    7     & End of restriction & circle, white, black  \\
\hline
    8     & Priority road       & diamond, white, yellow  \\
\hline
    9     & Give way       & triangle, white, red  \\
\hline
\end{tabular}
\end{center}
\end{table}

\section{Methods}

\subsection{Classifier}
As far as we don't aim in this experiment to built every neural network from scratch we took Classifier architecture from \cite{sermanet-ijcnn-11} as well as the idea to use a single grayscale channel instead of color images. Another reason is that the presented model achieves state-of-the-art using comparable simple and well-documented structure. \\
\\
The process of training machine learning model might be simplified to adjusting of parameters and hyperparameters to reach the point where loss function approaches global minimum. The number of parameters proportionally depends on the complexity of the task in which the model tends to solve. Another words vast number of parameters demand a considerable amount of training data.\\
\\
We can divide the training process into two stages: \\
\textbf{Stage 1}: Pretraining. We should pretrain the model using training dataset. Fortunately, from publicly accessible code repositories we can download already trained models.\\
\textbf{Stage 2}: Fine-tuning. Now we only have to retrain downloaded model using a decreased learning rate of 0.0001.

\begin{figure}[h]
  \centering
  \includegraphics[scale=0.4]{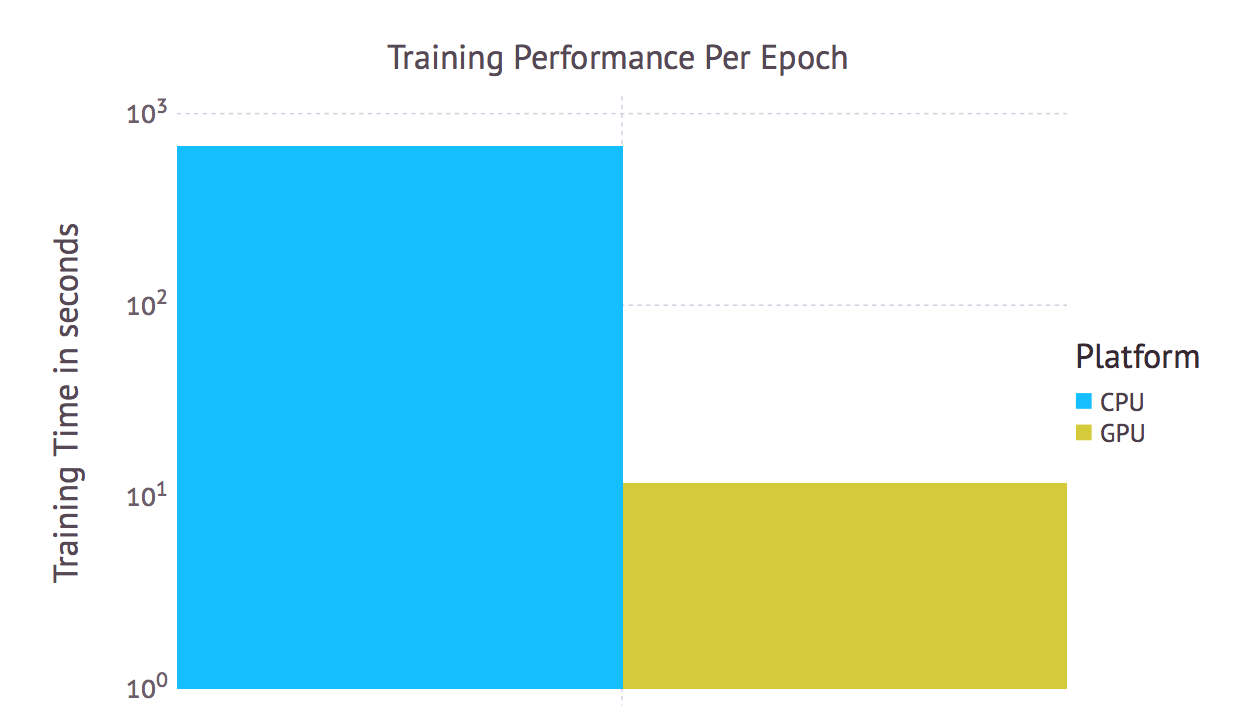}
  \caption{Performance comparison}
  \label{fig:cpugpu}
\end{figure}

\subsection{Classical Augmentation methods}
\begin{figure}[h]
  \centering
  \includegraphics[scale=0.33]{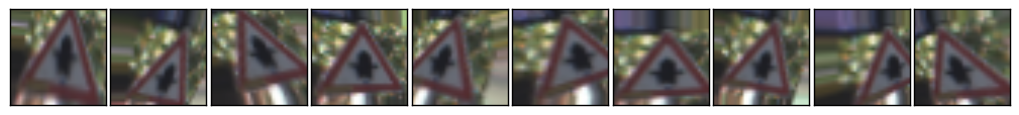}
  \includegraphics[scale=0.33]{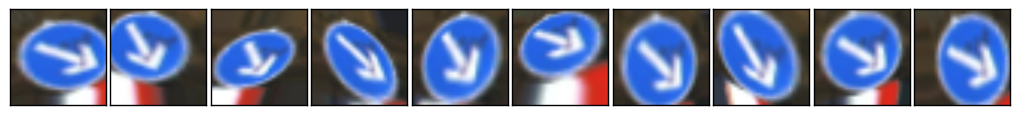}
  \caption{Examples of augmented images}
  \label{fig:fig1}
\end{figure}

In spite of all the data available in Superset, we wish it to have good diversity as desired sign images need to be present in varying sizes, lighting conditions and poses to be sure that our Classificator achieves the best possible result. To overcome this problem scientists generate extended dataset based on the original one. This methodology is known as data augmentation. Commonly used techniques to increase the amount of available data are Scaling, Translation, Rotation, Flipping, Adding Salt and Pepper noise, Lighting condition, Perspective transform and many other. Which slightly modify existing images. This might help to increment items in the dataset by thousands, however, those images won’t be as variant as we want them to be.
See an example of thee rotation method on Figure \ref{fig:fig1}.

\subsection{GAN}
The main contributor of this paper is Generative Adversarial Network, the idea of this neural network lays down in special structure composed of two networks which are trained simultaneously, improving attributes of each other. It is especially interesting for researchers because of its robustness to overfitting. Another benefit is that the training process doesn't require annotated training data. A generative model G captures the data distribution, and a discriminative model D estimates the probability that a sample came from the training data rather than G. The training
procedure for G is to maximize the probability of D making a mistake\cite{Goodfellow-GAN}.
Data type we are going to manipulate is represented as an image, from this point of view using advantages of Convolutional Neural Networks seems to be a great idea. However, there are few architecture guidelines  for stable Deep Convolutional GANs worked out in \cite{RadfordMC15}:
\begin{itemize}
\item Replace any pooling layers with strided convolutions  for discriminator and fractional-strided
convolutions in generator.
\item Use batchnorm in both the generator and the discriminator.
\item Remove fully connected hidden layers for deeper architectures.
\item Use ReLU activation in generator for all layers except for the output, which uses Tanh.
\item Use LeakyReLU activation in the discriminator for all layers.
\end{itemize}

\begin{figure}[h]
  \centering
  \includegraphics[scale=0.25]{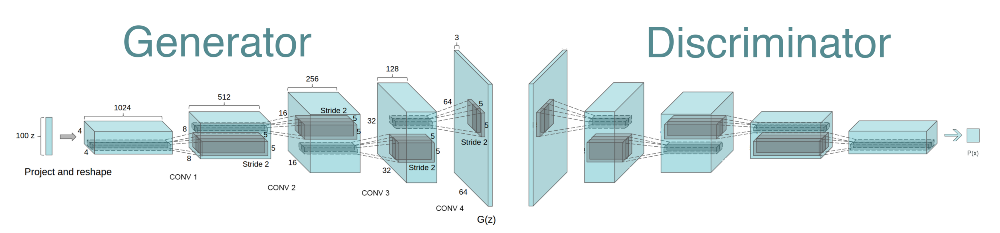}
  \caption{DCGAN Architecture}
  \label{fig:fig2}
\end{figure}

Using advises from our colleagues we end up with DCGAN\footnote{Deep Convolutional Generative Adversarial Network} architecture shown on 
Figure \ref{fig:fig2}.

Training process simplifies to launching from the command line a Python script and providing desired parameters, for example:

\begin{verbatim}
$ python3 main.py --dataset superset
--input_height=28 --output_height=28 --train
\end{verbatim}

The advantage of Using GAN for image generation is that it works for all sign classes, which is not true for data augmentation. For example, for “give way” sign, we can use a horizontal flip and slight rotation. For “Closed to all vehicles in both directions” all possible methods. Stop sign however can be flipped horizontally neither vertically, moreover, the rotation might cause problems with text detection if applies. Some mandatory signs after augmentation might totally change the meaning.

\begin{figure}[!htb]
\minipage{0.12\textwidth}
  \includegraphics[width=\linewidth]{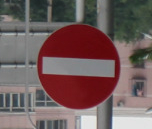}
\endminipage\hfill
\minipage{0.12\textwidth}
  \includegraphics[width=\linewidth]{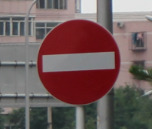}
\endminipage\hfill
\minipage{0.12\textwidth}
  \includegraphics[width=\linewidth]{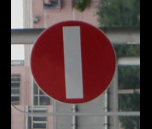}
\endminipage\hfill
\minipage{0.12\textwidth}
  \includegraphics[width=\linewidth]{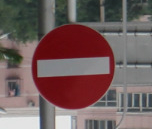}
\endminipage
\caption{No entry sign}
\end{figure}

\begin{figure}[!htb]
\minipage{0.12\textwidth}
  \includegraphics[width=\linewidth]{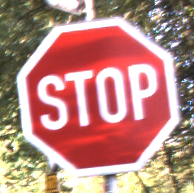}
\endminipage\hfill
\minipage{0.12\textwidth}
  \includegraphics[width=\linewidth]{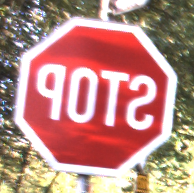}
\endminipage\hfill
\minipage{0.12\textwidth}
  \includegraphics[width=\linewidth]{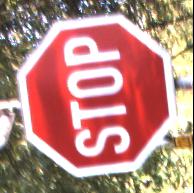}
\endminipage\hfill
\minipage{0.12\textwidth}
  \includegraphics[width=\linewidth]{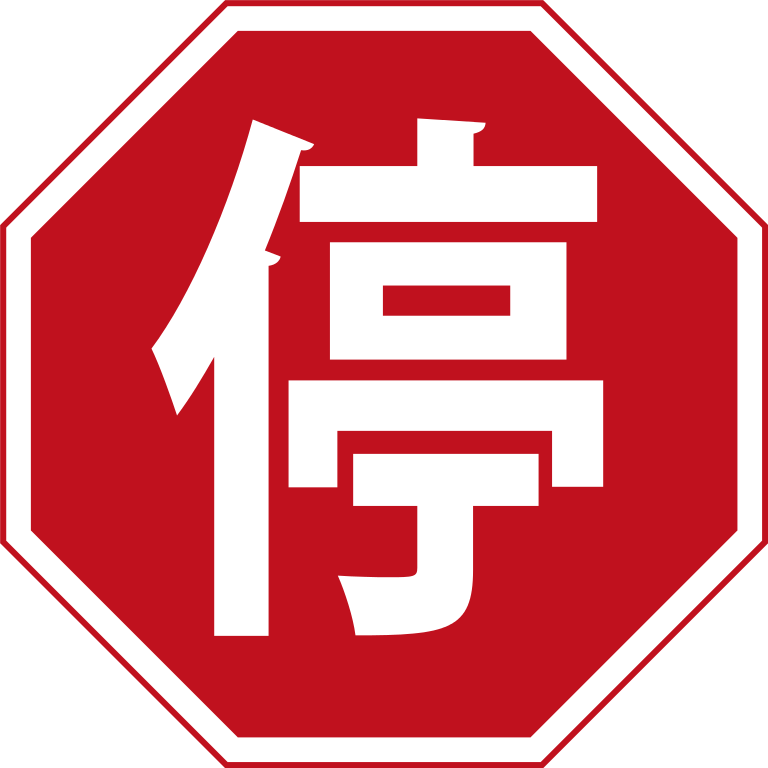}
\endminipage
\caption{Stop sign. Last image is Chinese sign}
\end{figure}

\subsubsection{Regularization}
By definition regularizations are techniques used to reduce the error by fitting a function appropriately on the given training set and avoid overfitting.\cite{CIS-215324} Other words, it is used for tuning the function by adding an additional penalty term in the error function. This is extremely important when the dataset is huge. $L(X,Y)$ is loss metrics, where $X$ is design matrix and $Y$ - observation vector (wages). In the way to penalize loss function, we add norm of weights vector $\omega$.\\
$L(X,Y) + \lambda N(\omega)$ where $\lambda$ is the tuning parameter that decides how much we want to penalize the flexibility of our model. $L$ function is multiplication of so-called LASSO\cite{DBLP:journals/corr/abs-1710-00598} and Ridge\cite{2018arXiv180510939K} norm. 
%\paragraph{Paragraph}

\section{Results}
In this section, we proceed to result which we receive during DCGAN training. Let's analyze graphs and numerical values we obtained during steps described in the introduction to understand if the hypothesis we claim matches the evidence. \\

First let's analyze pictures generated by our DCGAN from Figure \ref{fig:generated}. Images look noisy and blurry, however we simply able to recognize data encrypted in the sign. Training process took 25 epochs, this is enough to teach model to generate signs, however not enough to make output realistic on the level, that we as humans won't be able to identify which image is synthetic and which is real. Unfortunately for this experiment, we were limited by hardware, see Figure \ref{fig:cpugpu}\footnote{By Deepak Vinchhi, Co-Founder and Chief Operating Officer, Julia Computing, Inc.}, for this reason we apply the minimal satisfying amount of epochs for proof-of-concept. 
\\
\begin{figure}[!htb]
\minipage{0.12\textwidth}
  \includegraphics[width=\linewidth]{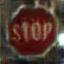}
\endminipage\hfill
\minipage{0.12\textwidth}
  \includegraphics[width=\linewidth]{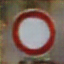}
\endminipage\hfill
\minipage{0.12\textwidth}
  \includegraphics[width=\linewidth]{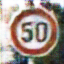}
\endminipage\hfill
\minipage{0.12\textwidth}
  \includegraphics[width=\linewidth]{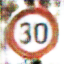}
\endminipage\hfill
\end{figure}

\begin{figure}[!htb]
\minipage{0.12\textwidth}
  \includegraphics[width=\linewidth]{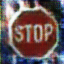}
\endminipage\hfill
\minipage{0.12\textwidth}
  \includegraphics[width=\linewidth]{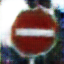}
\endminipage\hfill
\minipage{0.12\textwidth}
  \includegraphics[width=\linewidth]{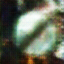}
\endminipage\hfill
\minipage{0.12\textwidth}
  \includegraphics[width=\linewidth]{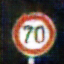}
\endminipage
\caption{Signs generated based on Superset}
  \label{fig:generated}
\end{figure}

\begin{figure}[!htb]
\minipage{0.25\textwidth}
\begin{tikzpicture}
\begin{axis}[
    title={Classifier testing on Test dataset},
    xlabel={Image Classes},
    ylabel={Accuracy},
    xmin=0, xmax=10,
    ymin=0, ymax=100,
    xtick={0,1,2,3,4,5,6,7,8,9},
    ytick={0,20,40,60,80,100},
    legend pos=south west,
    ymajorgrids=true,
    grid style=dashed,
]
 
\addplot[
    color=blue,
    mark=square,
    ]
    coordinates {
    (0,73.1)(1,77.5)(2,78)(3,78.8)(4,80.6)(5,75.8)(6,77.4)(7,79)(8,78.6)(9,77.3)
    };
    \addlegendentry{GTSRB trained}
\addplot[
    color=red,
    mark=square,
    ]
    coordinates {
    (0,75.2)(1,77.8)(2,78.4)(3,85.8)(4,81.8)(5,79.7)(6,83)(7,79)(8,82.5)(9,82.1)
    };
    \addlegendentry{GTSRB extended}
 
\end{axis}
\end{tikzpicture}
\endminipage

Our Classifier we trained on GAN-generated images perform on \textit{87.42\%} accuracy. Received results successively prove our beginning statements, however, we still need to analyze and preprocess data. Although we know how to improve our dataset using DCGAN, up to the present time we have to upgrade phase of data preparation.

\minipage{0.25\textwidth}
\begin{tikzpicture}
\begin{axis}[
    title={Improved Classifier testing on Test dataset},
    xlabel={Image Classes},
    ylabel={Accuracy},
    xmin=0, xmax=10,
    ymin=0, ymax=100,
    xtick={0,1,2,3,4,5,6,7,8,9},
    ytick={0,20,40,60,80,100},
    legend pos=south west,
    ymajorgrids=true,
    grid style=dashed,
]
 
\addplot[
    color=green,
    mark=square,
    ]
    coordinates {
    (0,84.9)(1,85.5)(2,84.9)(3,88.2)(4,89.5)(5,88.5)(6,90.3)(7,88.2)(8,86.9)(9,87.3)
    };
    \addlegendentry{trained on synthetic data}
\end{axis}
\end{tikzpicture}
\endminipage
  \caption{Test results}
  \label{fig:result}
\end{figure}
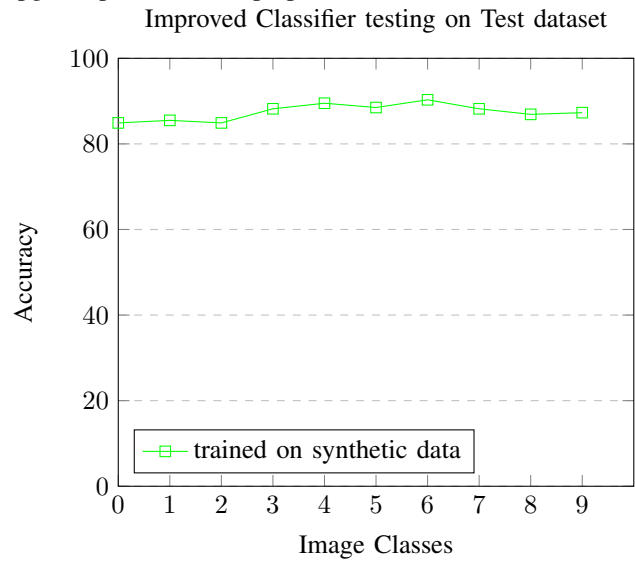

On Figure \ref{fig:result} we can see improvement which allows achieving augmentation, it's notable, but not significant. The average accuracy of pre-trained classifier showed \textit{77.61\%}, extended by augmented images increased this performance by 3\%, which is a good result when we are speaking about the neural network model.
\\

\section{Discussion and Future development}
When it comes to explaining the trivial task to an Artificial Intelligence system we often face difficulties. For example, next improvement step in traffic sign recognition we might propose an automated decision which classes of signs can be extracted from the dataset and augmented without losing meaning or accuracy, which method can be applied to the chosen class and which can not. This could be done using a simple decision tree.\\
To maximize proficiency of our Classifier we might extend or Superset not only with synthetically generated images but also apply augmentation to these images. For Generator we plan to use one of the regularization methods called cross-validation to achieve more realistic images on GAN output. Finally, the best crash test for this idea is real-time sign classification on the road.

\addtolength{\textheight}{-12cm}   % This command serves to balance the column lengths
                                  % on the last page of the document manually. It shortens
                                  % the textheight of the last page by a suitable amount.
                                  % This command does not take effect until the next page
                                  % so it should come on the page before the last. Make
                                  % sure that you do not shorten the textheight too much.

%%%%%%%%%%%%%%%%%%%%%%%%%%%%%%%%%%%%%%%%%%%%%%%%%%%%%%%%%%%%%%%%%%%%%%%%%%%%%%%%

%%%%%%%%%%%%%%%%%%%%%%%%%%%%%%%%%%%%%%%%%%%%%%%%%%%%%%%%%%%%%%%%%%%%%%%%%%%%%%%%

%%%%%%%%%%%%%%%%%%%%%%%%%%%%%%%%%%%%%%%%%%%%%%%%%%%%%%%%%%%%%%%%%%%%%%%%%%%%%%%%
%%%%%%%%%%%%%%%%%%%%%%%%%%%%%%%%%%%%%%%%%%%%%%%%%%%%%%%%%%%%%%%%%%%%%%%%%%%%%%%%


\begin{thebibliography}{1}

\bibitem{Goodfellow-GAN}
I.~Goodfellow, J.~Pouget-Abadie, M.~Mirza, B.~Xu, D.~Warde-Farley, S.~Ozair,
  A.~Courville, and Y.~Bengio.
\newblock Generative adversarial networks.
\newblock {\em arXiv preprint arXiv:1406.2661}, 2014.

\bibitem{2018arXiv180510939K}
D.~{Kobak}, J.~{Lomond}, and B.~{Sanchez}.
\newblock {Implicit ridge regularization provided by the minimum-norm least
  squares estimator when $n\ll p$}.
\newblock {\em arXiv e-prints}, May 2018.

\bibitem{RadfordMC15}
A.~Radford, L.~Metz, and S.~Chintala.
\newblock Unsupervised representation learning with deep convolutional
  generative adversarial networks.
\newblock {\em arXiv preprint arXiv:1511.06434}, 2015.

\bibitem{CIS-215324}
N.~Radford and C.~M. Bishop.
\newblock Pattern recognition and machine learning.
\newblock {\em Technometrics}, 49(3):366--366, 2007.

\bibitem{DBLP:journals/corr/abs-1710-00598}
A.~H. Ribeiro and L.~A. Aguirre.
\newblock Lasso regularization paths for narmax models via coordinate descent.
\newblock {\em arXiv preprint arXiv:1710.00598}, 2017.

\bibitem{sermanet-ijcnn-11}
P.~Sermanet and Y.~LeCun.
\newblock Traffic sign recognition with multi-scale convolutional networks.
\newblock In {\em Proceedings of International Joint Conference on Neural
  Networks (IJCNN'11)}, 2011.

\bibitem{Stallkamp2012}
J.~Stallkamp, M.~Schlipsing, J.~Salmen, and C.~Igel.
\newblock Man vs. computer: Benchmarking machine learning algorithms for
  traffic sign recognition.
\newblock {\em Neural Networks}, 2012.

\bibitem{Timofte-WACV-2009}
R.~Timofte, K.~Zimmermann, and L.~van Gool.
\newblock Multi-view traffic sign detection, recognition, and 3d localisation.
\newblock In {\em Ninth IEEE Computer Society Workshop on Application of
  Computer Vision}, pages 1--8, Snowbird, Utah, USA, December 2009.

\end{thebibliography}
\end{document}